%% file: main.tex
\documentclass[10pt,twocolumn,letterpaper]{article}

\usepackage[accsupp]{axessibility} 

 \usepackage{cvpr}              


\definecolor{cvprblue}{rgb}{0.21,0.49,0.74}
\usepackage[pagebackref,breaklinks,colorlinks,allcolors=cvprblue]{hyperref}
\usepackage{multirow}
\def\paperID{16} 
\def\confName{CVPR DG Workshop}
\def\confYear{2025}

\title{Mixture-of-Shape-Experts (MoSE): End-to-End Shape Dictionary Framework to Prompt SAM for Generalizable Medical Segmentation}

\author{Jia Wei$^{1,2}$, Xiaoqi Zhao$^1,2$, Jonghye Woo$^3$, Jinsong Ouyang$^1$, Georges El Fakhri$^{1,2}$,\\ Qingyu Chen$^2$, Xiaofeng Liu$^{1,2}$\\~\\
$^1$Dept. of Radiology and Biomedical Imaging, Yale University, New Haven, USA\\
$^2$Dept. of Biomedical Informatics and Data Science, Yale University, New Haven, USA\\
$^3$Dept. of Radiology, Massachusetts General Hospital and Harvard Medical School, Boston, USA\\
}

\begin{document}
\maketitle
\input{sec/0_abstract}    
\input{sec/1_intro}

\input{sec/2_formatting}

{
    \small
    \bibliographystyle{ieeenat_fullname}
    \bibliography{main}
}


\end{document}

%% file: sec/0_abstract.tex
\begin{abstract}
Single domain generalization (SDG) has recently attracted growing attention in medical image segmentation. One promising strategy for SDG is to leverage consistent semantic shape priors across different imaging protocols, scanner vendors, and clinical sites. However, existing dictionary learning methods that encode shape priors often suffer from limited representational power with a small set of offline computed shape elements, or overfitting when the dictionary size grows. Moreover, they are not readily compatible with large foundation models such as the Segment Anything Model (SAM). In this paper, we propose a novel \emph{Mixture-of-Shape-Experts (MoSE)} framework that seamlessly integrates the idea of mixture-of-experts (MoE) training into dictionary learning to efficiently capture diverse and robust shape priors. Our method conceptualizes each dictionary atom as a “shape expert,” which specializes in encoding distinct semantic shape information. A gating network dynamically fuses these shape experts into a robust shape map, with sparse activation guided by SAM encoding to prevent overfitting. We further provide this shape map as a prompt to SAM, utilizing the powerful generalization capability of SAM through bidirectional integration. All modules, including the shape dictionary, are trained in an end-to-end manner. Extensive experiments on multiple public datasets demonstrate its effectiveness. 
\end{abstract}

%% file: sec/1_intro.tex
\section{Introduction}
\label{sec:intro}

Accurate delineation of lesions or anatomical structures is a fundamental step in clinical diagnosis, intervention, and treatment planning~\cite{huang2023source,liu2022variational,liu2021segmentation,liu2022self}. While deep learning methods have recently demonstrated proficiency in segmentation, they face challenges in generalizing well across diverse domains, such as different clinical sites, scanner vendors, and imaging parameters~\cite{liu2023incremental,huang2023vicinal,liu2023memory,liu2022act}. These models often experience a substantial performance drop when evaluated on unseen out-of-domain data, limiting their practical applicability. To address this, domain generalization (DG) methods~\cite{liu2022deep} have been proposed, aiming to enable models to generalize across unseen domains. These methods typically focus on multi-source DG protocol relying on access to multiple source domain data, which can often be impractical due to privacy concerns and the complexity of data sharing in the clinical routine.

Instead, single domain generalization (SDG)~\cite{danish2024perturbing,desam,ccsdg,shape_dict,dapsam,dnorm} has garnered increasing attention as it trains exclusively on a single source domain yet aims to maintain robust performance on diverse target domains. Recent studies~\cite{chen2024learning,shape_dict,zhang2024shan} have highlighted that semantic shape information remains consistent across different medical domains, making it a valuable prior for SDG. For example,~\cite{shape_dict} further integrated dictionary learning to capture semantic shape priors, which tries to represent each segmentation mask as a linear combination of a set of basis elements (or atoms) in a shape dictionary by solving an iterative optimization problem. However, it is highly sensitive to the dictionary size to achieve a fragile balance (e.g., Fig.~\ref{hyper}(a)). When applied to medical SDG segmentation, it is challenging for a fixed small source domain dictionary to capture the full diversity of shapes across unseen domains. While increasing the dictionary size to encompass a wider range of shape variations can introduce the risk of overfitting, and cause the model to memorize specific features rather than generalize effectively across domains. Additionally, the shape dictionary in~\cite{shape_dict} is defined independently from the segmentation in two stages and be suboptimal.


In addition, the recently flourished large foundation model~\cite{han2024light,Liu2025vlm,han2024fair}, e.g., Segment Anything Model (SAM)~\cite{kirillov2023segment}, has shown powerful generalization ability. As a promptable foundation model in natural image segmentation, SAM has demonstrated remarkable performance in many zero-shot scenarios. Though it still struggles with specific medical segmentation due to the large inconsistency of natural and medical data~\cite{zhao2024inspiring,liu2024point}, and requires a mask or point prompt to indicate the to-be-segmented structure. In addition, how could dictionary learning be integrated into the SAM pipeline is underexplored. 








Motivated by these challenges, we propose a novel Mixture-of-Shape-Experts (MoSE) framework that integrates the advantages of mixture-of-experts (MoE) training with dictionary learning. The MoE scheme enables to store a diverse set of experts, sparsely selecting only a relevant subset with end-to-end networks based on the input~\cite{cai2024survey,shazeer2017outrageously,wang2022neural}. In our framework, each dictionary atom is conceptualized as a “shape expert” that encodes distinct semantic shape information derived from the source domain. Our learnable shape experts do not need to be offline computed before the segmentation training. A gating network dynamically generates sparse coding for the SAM embedding of each input image, selecting and combining the most relevant shape experts to form a robust, domain-invariant shape map. This sparse activation enables the model to capture a wider range of shape variations, mitigating the risk of overfitting. Furthermore, this shape map can be seamlessly incorporated back into SAM as a prompt, thereby guiding the segmentation process and markedly enhancing generalization across unseen domains. Our key contributions are:

$\bullet$ We introduce an end-to-end shape dictionary framework MoSE that employs MoE strategy to efficiently capture diverse and robust shape priors by a dynamic gating mechanism. It selectively fuses shape experts enforced by sparse activation to alleviate overfitting.


$\bullet$ A bidirectional integration scheme with the SAM pipeline, leveraging its powerful encoder for sparse coding, while providing the shape map as a prompt. It is regularized by utilization balancing and warm-up strategy.


$\bullet$ Unlike traditional dictionary learning methods, we can keep achieving performance gains by increasing the dictionary size (number of shape experts) from 256 to 1024. Extensive experiments demonstrate that our proposed MoSE substantially outperforms state-of-the-art SDG methods and the adapted SAM.

%% file: sec/2_formatting.tex
\section{Related Works}

\begin{figure*}[t]
\begin{center} 
\includegraphics[width=\linewidth]{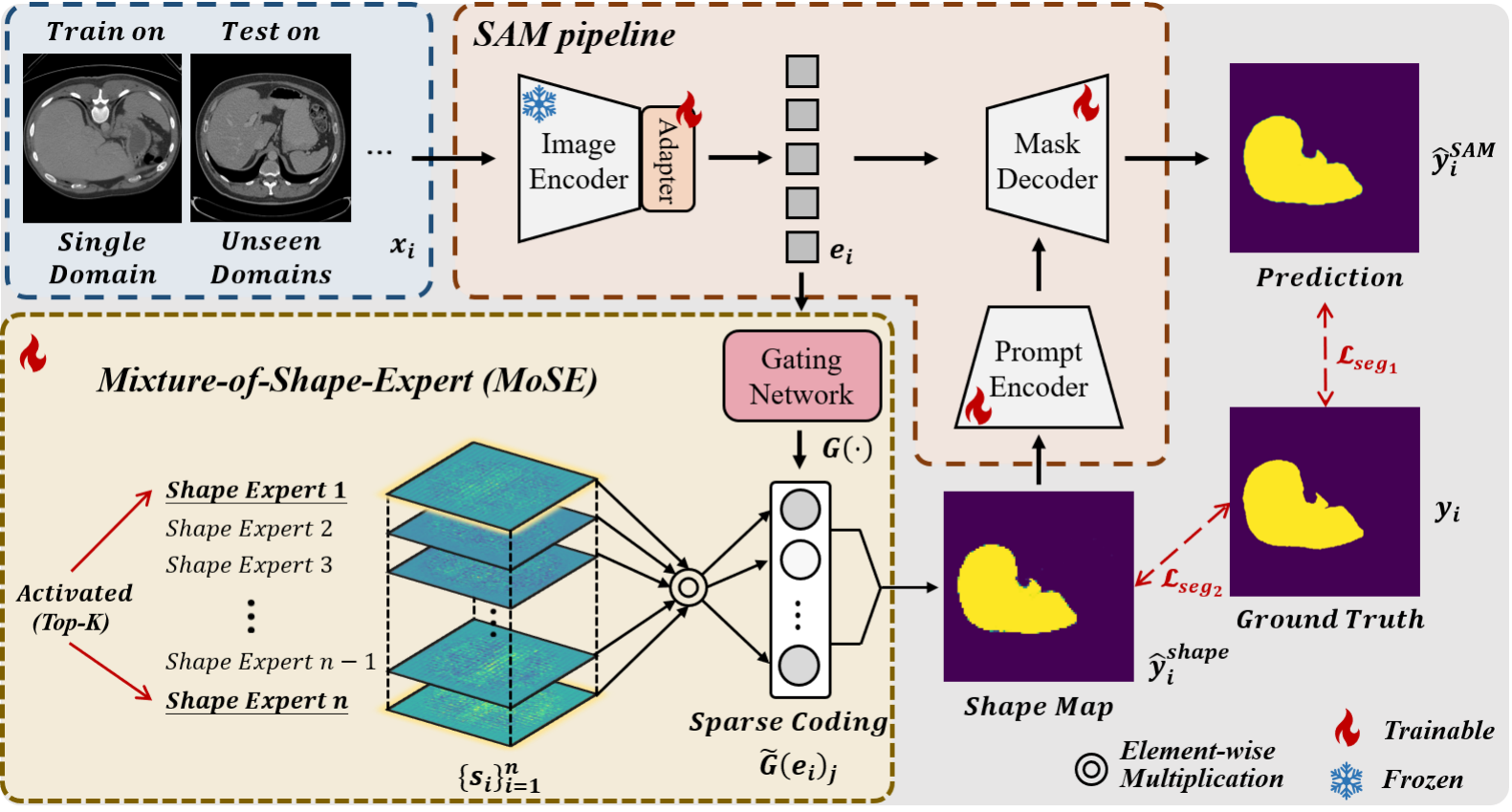}
\end{center} 
\caption{ 
Overview of the proposed Mixture-of-Shape-Experts (MoSE) framework for Single Domain Generalization (SDG). The framework leverages a dictionary of shape experts to store diverse shape priors, which are dynamically combined into a shape map via sparse coding generated by SAM encoder and gating network. The shape map serves as a prompt integrated into the SAM pipeline.
} 
\label{fig1}\end{figure*}

\subsection{Single Domain Generalization.}

Domain generalization (DG)~\cite{khan2021mode,chokuwa2023generalizing} aims to generalize a model trained on multiple source domains to a distributionally different target domain. Previous progress on DG is mainly made by enabling models to learn from multi-source domain data to reduce the model bias.

Recently, single domain generalization (SDG)~\cite{danish2024perturbing,desam,ccsdg,shape_dict,dapsam,dnorm} has attracted growing attention, where only one source domain is available during training and the model is evaluated on multiple unseen target domains. This can be particularly challenging due to variations in imaging protocols, equipment, and patient populations across different medical centers. There are several pioneer works on image classification~\cite{qiao2020learning,volpi2018generalizing,wang2021learning,fan2021adversarially,zhang2022exact} to improve the performance on unseen domains while training on a single source domain. In particular,~\cite{qiao2020learning,volpi2018generalizing,wang2021learning} propose to use a data augmentation scheme where diverse input images are generated via adversarial training. Instead, \cite{fan2021adversarially,zhang2022exact} propose to utilize the normalization mechanism to adapt the feature distribution to unseen domains. It also been explored for object detection~\cite{danish2024perturbing,wu2022single} . \cite{vidit2023clip} further proposed to leverage a self-supervised vision-language model to guide the training of an object detector.

For segmentation tasks,~\cite{ouyang2022causality,su2023rethinking} also adopt data augmentation approach to expose a segmentation model to synthesized domain-shifted training. Similarly,~\cite{xu2022adversarial} synthesizes the new domains via learning an adversarial domain synthesizer. In addition,~\cite{hu2023devil} used contrastive training  with shallower features of each image and its style-augmented counterpart. However, the effectiveness of SDG is directly influenced by the realism of generated data, which is also a long-lasting challenge. Recent studies~\cite{chen2024learning,shape_dict,zhang2024shan} have emphasized the consistency of semantic shape information across various medical domains, establishing it as a valuable prior for SDG. How to efficiently exploit this prior knowledge is a challenging task.

\subsection{Dictionary Learning} seeks to find a set of basic elements to compose a dictionary such that a given input
can be well represented by a sparse linear combination of
these learned elements~\cite{tovsic2011dictionary,liu2018joint}. Incorporating dictionary learning into medical image segmentation usually involves learning a set of basis elements (the dictionary) that can sparsely represent image patches~\cite{tong2015discriminative,zhao2021survey}. It has been applied to enhance segmentation performance by capturing essential structural features of medical images. For instance, combining dictionary learning with clustering algorithms has led to unsupervised adaptive segmentation methods, where dictionaries serve as clustering centers, and sparse representation is used for segmentation~\cite{zhao2021survey}. A notable approach of SDG segmentation is the integration of shape priors through dictionary learning~\cite{shape_dict}. By constructing a shape dictionary from the available source domain data, models can capture semantic shape priors that are invariant across domains. However, there is a trade-off of dictionary size and performance~\cite{shape_dict}. This raises an important question: How to increase the dictionary size to encompass a broader spectrum of shape priors, while simultaneously mitigating the risk of overfitting?

\subsection{Mixture of Experts} is founded on a straightforward yet effective concept: distinct model components, referred to as experts, are trained to specialize in different tasks or data characteristics~\cite{cai2024survey}. It is initially introduced in~\cite{jacobs1991adaptive,jordan1994hierarchical}, has undergone extensive exploration and advancement as in the era of foundation models~\cite{lo2024closer,mu2025comprehensive}. In MoE framework, only the most relevant experts are activated for a given input, optimizing computational efficiency while leveraging a diverse set of specialized knowledge. This flexible and scalable approach aligns with the scaling law, enabling greater model capacity without a proportional increase in computational costs. \cite{wang2024sam} proposes to combine different fine-tuned SAM models with MoE. In this work, we introduce a novel concept of Mixture-of-Shape-Experts (MoSE), to explore the efficiency of MoE. It is essentially different to the mixture of networks.

\section{Methodology}

In the SDG setting, we are provided with a single source domain dataset $D^s = {(x_i, y_i)}_{i=1}^N$, consisting of $N$ pairs of source domain input slices $x_i \in \mathbb{R}^{H \times W}$ and their corresponding segmentation masks $y_i \in \mathbb{R}^{H \times W \times C}$. Here, $H$, $W$, and $C$ represent the height, width, and segmentation class, respectively. After training on the source domain dataset, the model's performance is evaluated on unseen target domains $D^t = \{ D^t_1, D^t_2, \dots, D^t_n \}$. The overview of our proposed MoSE framework is presented in Fig.~\ref{fig1}.


\subsection{Mixture-of-Shape-Experts (MoSE) Training}

In traditional dictionary learning methods~\cite{shape_dict}, the dictionary is built offline, with fixed atoms that are combined with coefficients generated by a regression branch to produce shape priors. However, this approach has two key limitations in medical SDG segmentation: 1) A fixed, small source domain dictionary struggles to capture the diverse shapes present in unseen target domains, hindering generalization. 2) The offline construction of the dictionary is decoupled from the segmentation task, limiting its adaptability and effectiveness.

Inspired by the Mixture-of-Experts (MoE) training system \cite{shazeer2017outrageously}, which enables a significant increase in model capacity while maintaining a constant computational cost through sparse activation of experts, we propose integrating MoE into dictionary learning. In our approach, the atoms of the dictionary are conceptualized as "shape experts," each storing diverse shape priors learned from the source domain data. Specifically, we define $n$ learnable experts \cite{cai2018novel,conde2022model} of shape as $\{ s_i \}_{i=1}^n$, where each $s_i \in \mathcal{R}^{h \times w}$. To reduce parameter overhead and align with the size of the prompt encoder in SAM, we set $h = \frac{H}{4}$ and $w = \frac{W}{4}$.


We then utilize a lightweight convolutional neural network (CNN) as the gating network, which generates sparse coding to combine the shape experts. Specifically, for each input image \( x_i \) embedded into \( e_i \) with SAM image encoder, the gating network produces an output \( G(e_i) \in \mathbb{R}^{h \times w \times n} \), which assigns pixel-wise weights to the shape experts. To enforce sparsity and select only the most relevant experts, we apply a \textbf{Top-K selection} strategy based on absolute values, retaining the \( k \) shape experts with the highest responses at each pixel: 
\begin{equation}
\tilde{G}(e_i)_{j} = 
\begin{cases}  
G(e_i)_{j}, & \text{if } j \in \mathcal{T}(e_i), \\  
0, & \text{otherwise}
\end{cases}
\end{equation}

\noindent 
where \( \mathcal{T}(e_i) \) denotes the set of indices corresponding to the \textbf{top \( k \) largest absolute values} of \( G(e_i) \) at each pixel location. The final shape map is then computed as: 
\begin{equation}
\hat{y}_i^{\text{shape}} = \sum_{j=1}^n \tilde{G}(e_i)_{j} \odot s_j,
\end{equation}

\noindent where \( \tilde{G}(e_i)_{j} \) represents the sparsified gating weight map for the \( j \)-th shape expert, $\odot$ indicates element-wise multiplication broadcast across the spatial dimension. Intuitively, each pixel $p$ in $\hat{y}_i^{\mathrm{shape}}$ is formed by a linear combination of the top-$k$ experts deemed most relevant by the gating network. Additionally, the generated shape map $\hat{y}_i^{\mathrm{shape}}$, formed through this pixel-wise combination, is then combined with the label $y_i$ to compare the segmentation loss, enabling end-to-end training, as detailed in Section~\ref{seg_loss}.

\subsection{Integrating Shape Map into SAM Pipeline}

Our MoSE seamlessly incorporates SAM with a bidirectional scheme to fully exploit its generalizability. In addition to assisting with sparse coding, inspired by prompt learning~\cite{jia2022visual,li2024mamba,li2024text,wei2024fnet}, we incorporate the shape map \( y_i^{\text{shape}} \) as a prompt, enabling it to interact with the image embedding. This integration introduces prior shape information, guiding the segmentation process and enhancing the model's generalization to unseen domains.  


Specifically, the shape map \( \hat{y}_i^{\text{shape}} \) is first passed through a sigmoid activation function to normalize its values within \([0,1]\), ensuring compatibility with the prompt representation: 
\begin{equation}
    y_i^{\text{prompt}} = \sigma(\hat{y}_i^{\text{shape}}),
\end{equation}

\noindent 
where \( \sigma(\cdot) \) denotes the sigmoid function. The processed shape prompt \( y_i^{\text{prompt}} \) is then integrated into the SAM framework \( f_{\text{SAM}}(\cdot) \), interacting with the image embedding to guide the mask decoder. This results in the final segmentation prediction: 
\begin{equation}
    \hat{y}_i^{\text{SAM}} = f_{\text{SAM}}(x_i, y_i^{\text{prompt}}).
\end{equation}

\begin{table*}[t!]
\centering
\resizebox{1.35\columnwidth}{!}{%
\begin{tabular}{l|l|c|c}
\hline\hline
\textbf{Datasets} & \textbf{Modality} & \textbf{\# of Cases} & \textbf{\# of Slices (with Ground Truth)} \\ \hline\hline
\textbf{BTCV}~\cite{btcv}                       & CT                & 30                   & 1542                                       \\ \hline
\textbf{LITS}~\cite{lits}                       & CT                & 131                  & 19160                                      \\ \hline
\textbf{CHAOS}~\cite{kavur2021chaos}                      & CT                & 20                   & 2341                                       \\ \hline
\textbf{AMOS (Evaluation Set)}~\cite{ji2022amos}      & CT                & 100                  & 4865                                       \\ \hline
\textbf{WORD (Evaluation Set)}~\cite{luo2022word}      & CT                & 20                   & 1110                                       \\ \hline\hline
\end{tabular}%
}
\caption{Overview of liver CT datasets used in this study.}
\label{tab:datasets}
\end{table*}

\begin{figure*}[!t]
\begin{center}
\includegraphics[width=1\linewidth]{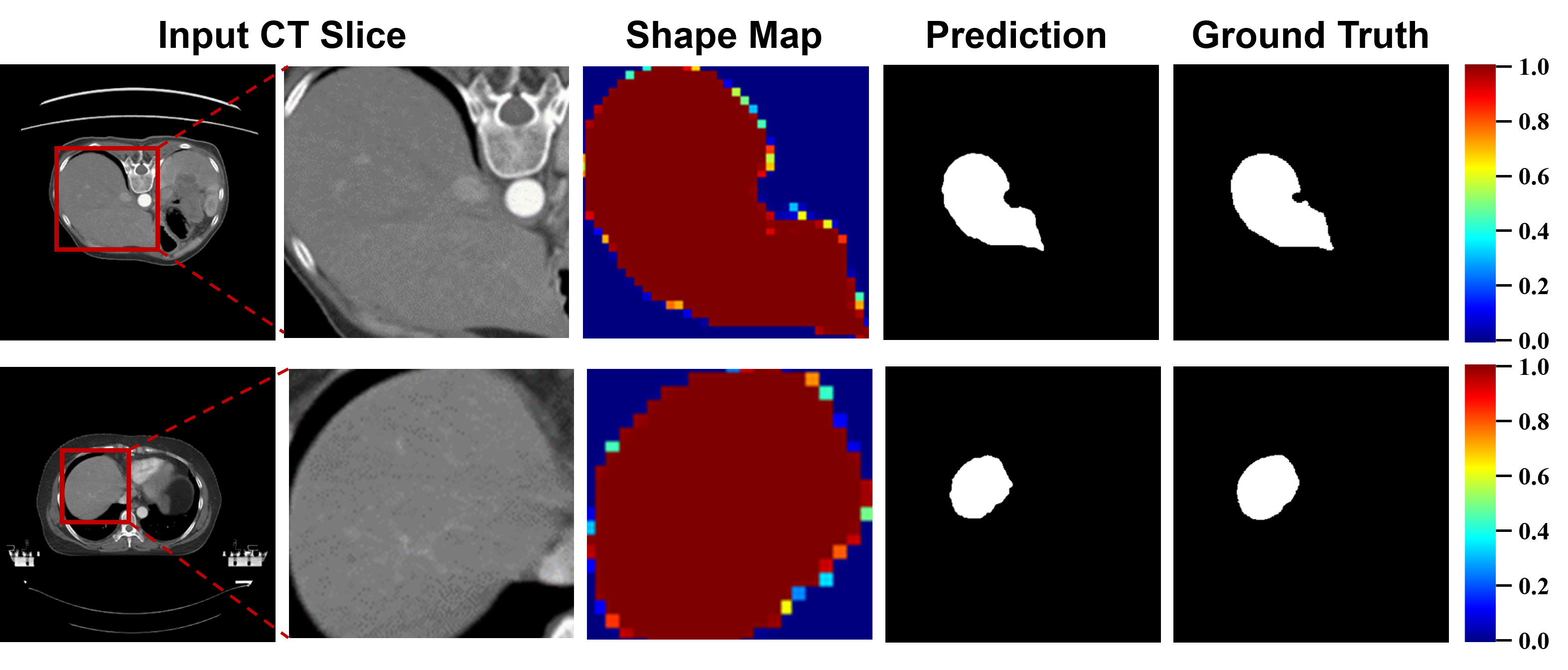}
\end{center}  
\caption{Visualization of the shape maps and segmentation results. The shape map is a heatmap of shape priors formed by combining shape experts and coefficients generated by the gating network. It is then used as a prompt in the subsequent SAM prompt encoder to assist in segmentation.}
\label{seg_results}
\end{figure*}

\subsection{Training Objectives and Strategy}

\subsubsection{Segmentation Loss} \label{seg_loss}
Both the final segmentation prediction \( \hat{y}_i^{\text{SAM}} \) and the intermediate shape map output \( \hat{y}_i^{\text{shape}} \) are supervised using the label \( y_i \). Following SAMed~\cite{samed}, we employ a composite segmentation loss that combines the Cross-Entropy (CE) loss and the Dice loss to train the network modules and shape dictionary on the single source domain, which is defined as: 
\begin{equation} \label{seg}
    \mathcal{L}_{\mathrm{seg}}(\hat{y}, y_i) = (1 - \lambda) \cdot \mathcal{L}_{\text{CE}}(\hat{y}, y_i) + \lambda \cdot \mathcal{L}_{\text{Dice}}(\hat{y}, y_i),
\end{equation}

\noindent where \( \lambda \) is a trade-off factor. \( \hat{y} \) denotes either \( \hat{y}_i^{\text{SAM}} \) or \( \hat{y}_i^{\text{shape}} \) in $\mathcal{L}_{seg1}$ or $\mathcal{L}_{seg2}$.

\subsubsection{Utilization Balancing and Warm-Up Strategy in MoSE Training}

The gating network in the Mixture-of-Shape-Experts (MoSE) framework tends to reinforce a self-imbalanced state, assigning high weights to a limited number of shape experts. To encourage balanced usage, we introduce a Coefficient of Variation (CV) regularization~\cite{bengio2015conditional,shazeer2017outrageously} on the sparse coding: 
\begin{align} 
&\mathcal{P}_{\text{CV}} \left( \boldsymbol{\alpha}^{(i)}, \dots, \boldsymbol{\alpha}^{(T)} \right) = \frac{\operatorname{Var}(\bar{\boldsymbol{\alpha}})}{\left(\sum_{i=1}^{n} \bar{\alpha}_i / n \right)^2}, \\&\text{where}~ \bar{\boldsymbol{\alpha}} = \sum_{i=1}^{T} \boldsymbol{\alpha}^{(i)}.
\end{align}

Minimizing this term encourages experts to be invoked more evenly. Though computing this regularization over the entire dataset is infeasible, we estimate and minimize it at the batch level.

Additionally, hard sparsification of the gating network may halt gradient backpropagation, leading to stagnation at its initial state. To address this, we propose to delay hard thresholding and initially train the MoSE layer using an \( \ell_1 \)-norm penalty \begin{equation}\mathcal{P}_{\ell_1} = \sum_{i=1}^{T} \left\Vert \boldsymbol{\alpha}^{(i)} \right\Vert_1\end{equation} for several iterations following~\cite{wang2022neural}. After that, we enable Top-K selection to enforce stricter sparsity in $G(e_i)$.

\subsubsection{Overall Loss}

Putting all terms together, our total training loss is 
\begin{align}
\label{eq:overall_loss}
\mathcal{L} 
=& \sum_{i=1}^N
\Bigl[
   \mathcal{L}_{\mathrm{seg}}(\hat{y}_i^{\mathrm{SAM}}, y_i)
 + \mathcal{L}_{\mathrm{seg}}(\hat{y}_i^{\mathrm{shape}}, y_i)
\Bigr]
\\&+ \beta \cdot \begin{cases}
\mathcal{P}_{\ell_1}, & \text{if iteration} \leq T_{\text{warm-up}}, \\
\mathcal{P}_{\mathrm{CV}}, & \text{otherwise}.
\end{cases}
\end{align}
where \( \beta \) controls the relative importance of either the \( \ell_1 \) warm-up penalty or the CV regularization, depending on the iteration number \( T_{\text{warm-up}} \).

\subsection{Inference in Unseen Domains}

Once trained on $D^s$, our MoSE with SAM model can process an unseen target-domain image $x^{t}$ without requiring additional adaptation to $D^t$, by computing: 
\begin{align}
\hat{y}^{\,\mathrm{shape}} = \sum_{j=1}^n \tilde{G}(e^t)_{j} \odot s_j, 
\\
\hat{y}^{\,\mathrm{SAM}} = f_{\text{SAM}}\bigl(x^t, \sigma(\hat{y}^{\,\mathrm{shape}})\bigr).
\end{align}

\noindent Here, $e^t$ is the target embedding generated from $x^t$. The final segmentation mask is $\hat{y}^{\,\mathrm{SAM}}$, which benefits from shape priors encapsulated in the dictionary.

\begin{table*}[!t]
\centering
\caption{Quantitative comparison and ablation study of SDG (trained on BTCV) results on unseen B-E domains. The best results are in \textbf{bold}. All results are reported as mean$\pm$standard deviation, based on 3 random seeds.}
\renewcommand{\arraystretch}{1.3}
\resizebox{1\linewidth}{!}{
\begin{tabular}{l|ccccc|ccccc}
\hline \hline
\multirow{2}{*}{Method} & \multicolumn{5}{c|}{\textbf{Dice Coefficient (Dice, mean$\pm$std)} $\uparrow$} & \multicolumn{5}{c}{\textbf{Hausdorff Distance (HD, mean$\pm$std)} $\downarrow$} \\ \cline{2-11}
                        & B    & C    & D    & E    & Avg.  & B    & C    & D    & E    & Avg.  \\ \hline  \hline

D-Norm~\cite{dnorm}    & 69.6$\pm$5.2  & 70.5$\pm$7.1  & 70.5$\pm$3.6  & 73.6$\pm$8.9  & 71.0$\pm$3.5  & 84.4$\pm$9.1  & 80.6$\pm$21.7  & 54.7$\pm$14.6  & 83.5$\pm$19.0  & 75.8$\pm$9.9  \\

CCSDG~\cite{ccsdg}     & 91.2$\pm$0.9  & 88.2$\pm$0.2  & 87.4$\pm$0.1  & 86.4$\pm$1.9  & 88.4$\pm$0.3  & 36.0$\pm$2.6  & 35.3$\pm$3.8  & 16.4$\pm$0.7  & 10.0$\pm$0.3  & 24.4$\pm$0.7  \\

SAMed~\cite{samed}    & 88.2$\pm$0.5  & 78.2$\pm$2.7  & 88.6$\pm$0.1  & 86.5$\pm$0.2  & 85.4$\pm$0.7  & 27.0$\pm$1.9  & 40.3$\pm$1.7  & 11.6$\pm$2.3  & \textbf{9.5$\pm$0.4}   & 22.1$\pm$1.2  \\

DeSAM-B~\cite{desam}    & 79.8$\pm$0.6  & 88.5$\pm$0.9  & 89.2$\pm$0.7  & 80.0$\pm$3.5  & 84.2$\pm$1.2  & \textbf{19.6$\pm$2.1}  & 29.9$\pm$3.2  & 18.5$\pm$0.1  & 25.9$\pm$14.5  & 23.5$\pm$6.2  \\

DeSAM-P~\cite{desam}    & 80.0$\pm$0.3  & 86.9$\pm$0.2  & 87.2$\pm$1.9  & 78.0$\pm$3.1  & 83.2$\pm$0.6  & 20.1$\pm$3.5  & 34.2$\pm$2.7  & 19.9$\pm$1.8  & 37.3$\pm$20.4  & 28.4$\pm$6.7  \\

DAPSAM~\cite{dapsam}  & 91.1$\pm$1.0  & 86.6$\pm$1.3  & 86.5$\pm$1.0  & 89.3$\pm$1.4  & 88.4$\pm$0.7  & 26.3$\pm$2.6  & 39.7$\pm$0.6  & 22.8$\pm$1.8  & 21.5$\pm$5.7  & 27.6$\pm$2.6  \\

\hline

Baseline & 91.0$\pm$0.7  & 84.9$\pm$2.5  & 87.2$\pm$0.9  & 86.0$\pm$5.7  & 87.3$\pm$2.2  & 30.6$\pm$4.3  & 39.8$\pm$4.8  & 15.4$\pm$3.4  & 14.1$\pm$3.5  & 25.0$\pm$3.0  \\\hline

MoSE w/o MoE & 90.4$\pm$2.0 & 85.3$\pm$4.5 & 89.5$\pm$1.5 & 88.9$\pm$0.8 & 88.5$\pm$2.3 & 38.9$\pm$9.3 & 40.5$\pm$8.7 & 13.8$\pm$3.5 & 14.5$\pm$3.4 & 25.3$\pm$5.1  \\
MoSE (\textit{Ours})  & \textbf{92.7$\pm$0.2}  & \textbf{90.3$\pm$0.7}  & \textbf{91.6$\pm$0.7}  & \textbf{90.9$\pm$0.3}  & \textbf{91.4$\pm$0.5}  & 20.7$\pm$2.3  & \textbf{25.6$\pm$2.3}  & \textbf{9.7$\pm$1.5}  & 12.1$\pm$3.1  & \textbf{17.0$\pm$2.1}  \\ \hline  \hline
\end{tabular}} 
\label{tab1}
\end{table*}

\section{Experiments and Results}

\subsection{Datasets and Pre-processing} 

We collect liver CT scans from five publicly available datasets: \textbf{A)} BTCV~\cite{btcv} (1542 slices), \textbf{B)} LITS~\cite{lits} (19,160 slices), \textbf{C)} CHAOS~\cite{kavur2021chaos} (2341 slices), \textbf{D)} {evaluation set} of AMOS~\cite{ji2022amos} (4865 slices), and \textbf{E)} {evaluation set} of WORD~\cite{luo2022word} (1110 slices). 

All slices are resampled to a uniform axial resolution of \( 384 \times 384 \) pixels. We select the BTCV dataset as the source domain for training and evaluate the model's performance on the other four unseen domains. {For each domain of B-E, we use a 20/80 split for validation and testing.}

\begin{figure*}[!t]
\begin{center}
\includegraphics[width=1\linewidth]{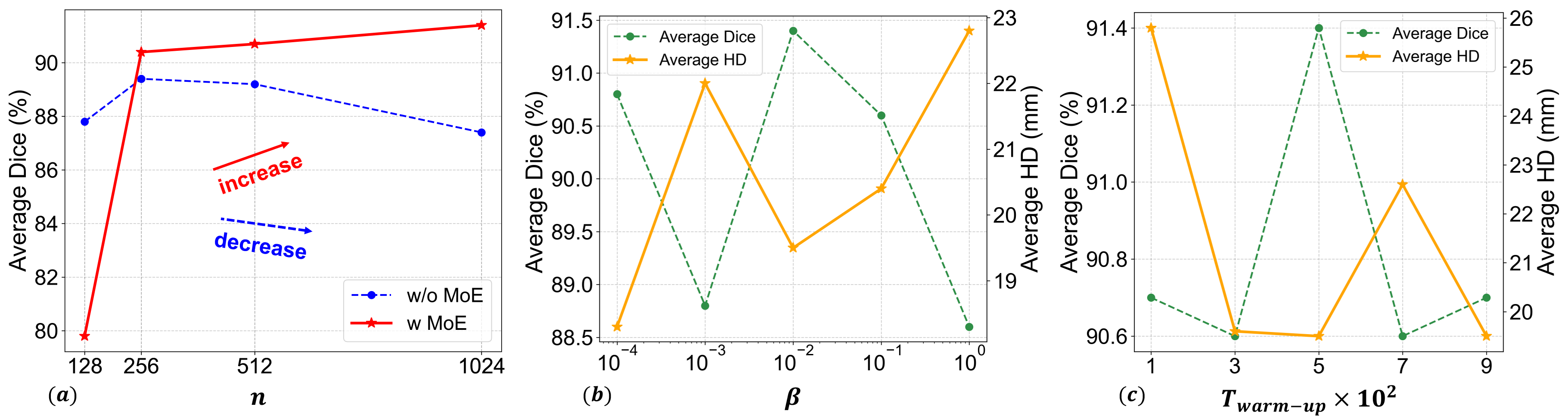} 
\end{center}
\caption{Sensitivity analysis of hyperparameters. (a) With or without MoE for different dictionary sizes $n$, where the number of selected experts is set as $k=\frac{n}{2}$ for each case. (b) and (c) Dice and HD of different (b) \( \beta \) and (c) \( T_{warm-up} \). All results are averaged across all target domains.} 
\label{hyper}
\end{figure*}

\begin{figure*}[t]
\begin{center} 
\includegraphics[width=0.95\linewidth]{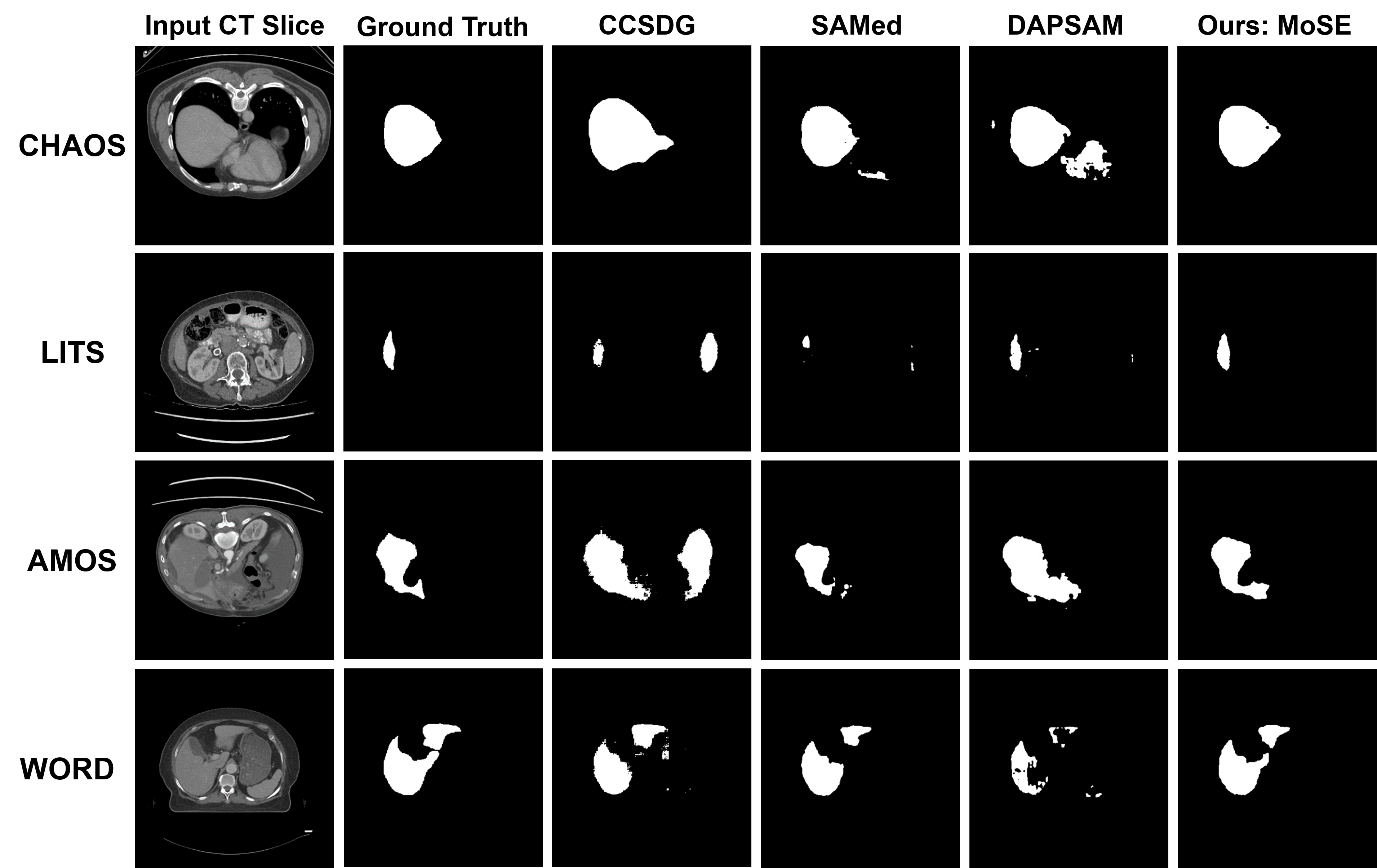}
\end{center}
\caption{ 
Visualization of segmentation results on unseen domains compared with other state-of-the-art SDG methods.
}
\label{fig2}
\end{figure*}


Subsequently, each volume is normalized by subtracting its minimum intensity value and dividing by its intensity range. The details of these datasets are shown in Table.~\ref{tab:datasets}. The datasets in this study all undergo preprocessing with consistent window width (WW) and window level (WL) values. The WW is set to 600, and the WL is generally set to 100 for most datasets following~\cite{li2021semantic}, except for the CHAOS dataset, which uses a higher WL of 1100, suitable for the liver. Additionally, for the LITS dataset, we merge the liver tumor class into the liver class following the same pre-processing adopted in~\cite{wen2024denoising}.


\begin{figure}[t] 
\begin{center} 
\includegraphics[width=1\linewidth]{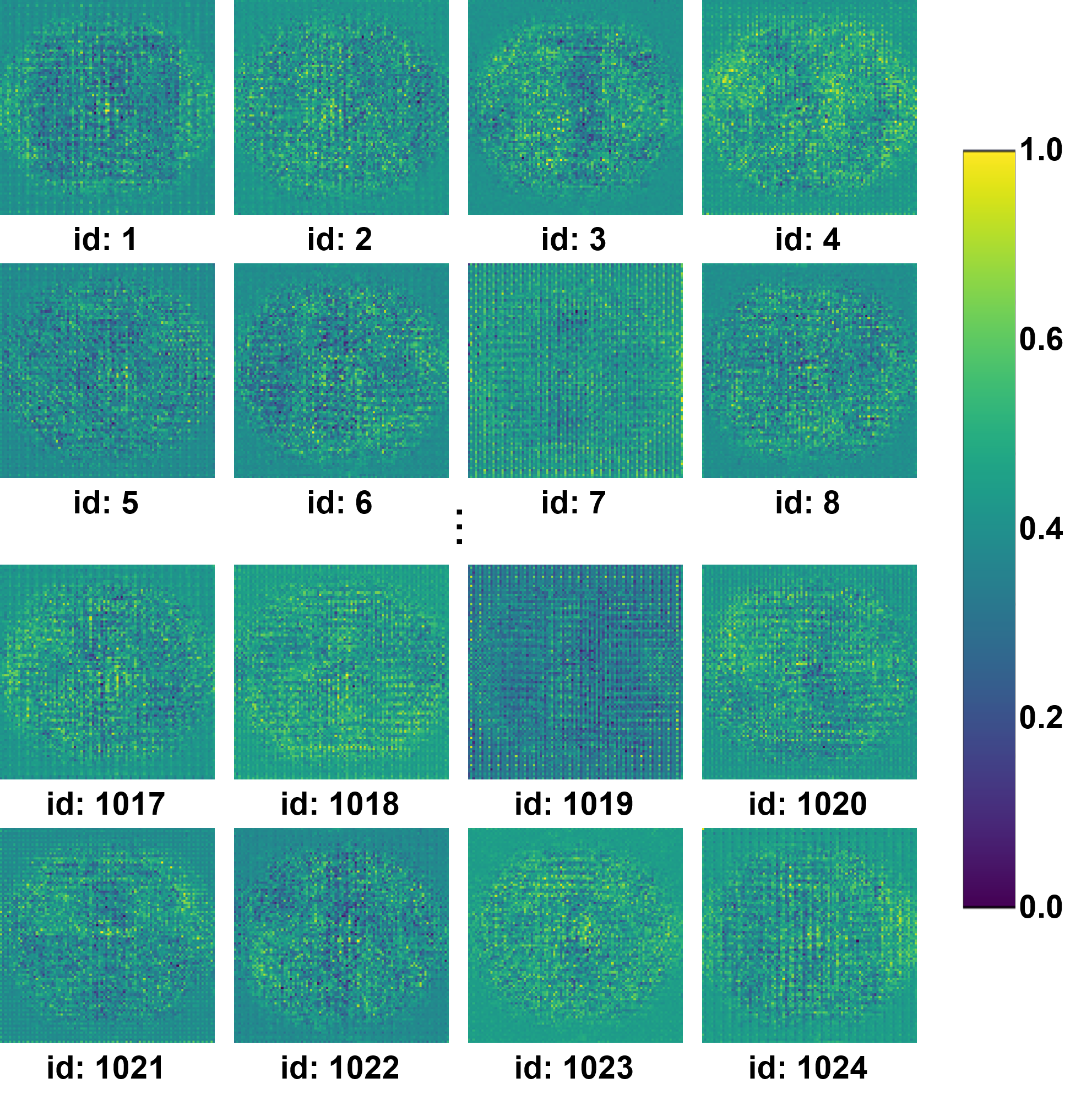}
\end{center}
\caption{Visualization of the learned shape experts.} 
\label{fig1}
\end{figure}

\subsection{Implementation Details}
We adopt the vanilla SAM~\cite{kirillov2023segment} and introduce two trainable MLP-structured adapters into each layer of the encode, serving as our \underline{\textit{baseline model}}, following the design proposed in~\cite{adaptformer}. {Based on the hyperparameter validation}, the total number of shape experts is set to \( n=1024 \) and the Top-K selection parameter is set to \( k=512 \). We further examine the impact of these settings in our ablation study. All experiments are conducted using the `ViT-B' version of SAM. The hyperparameters \( \beta \) and \( T_{\text{warm-up}} \) are set to \( 10^{-2} \) and \(5\times10^{2}\), respectively. We empirically set \( \lambda \) to 0.8 to balance the contribution of the two loss components.

The rank of the trainable MLP-structured adapters is set to 4, which facilitates a balance between computational efficiency and performance. We initialize training with a learning rate of \( 3 \times 10^{-3}\) and utilize the AdamW optimizer with a weight decay of 0.1. The training processes of all methods are capped at a maximum of 150 epochs for a fair comparison. Due to GPU memory limitations, we did not experiment with larger values of \( n \), though increasing \( n \) further may lead to even better results.

\subsection{Results of Single Domain Generalization}

The segmentation performance is quantitatively assessed using two widely-used Dice coefficient [\%] (Dice, higher is better) and Hausdorff Distance [mm] (HD, lower is better). In Table~\ref{tab1}, we compare our proposed MoSE with other state-of-the-art SDG methods, including CNN-based D-Norm~\cite{dnorm} and CCSDG~\cite{ccsdg}, as well as ViT-based DeSAM-B~\cite{desam}, DeSAM-P~\cite{desam}, and DAPSAM~\cite{dapsam}. 

MoSE achieves the highest average Dice of 91.4\% across all domains, surpassing other methods such as D-Norm (71.0\%), CCSDG (88.4\%), SAMed (85.4\%), and DAPSAM (88.4\%). In particular, MoSE excels in all individual target domains, achieving the highest Dice scores of 92.7\% on LITS (B), 90.3\% on CHAOS (C), 91.6\% on AMOS (D), and 90.9\% on WORD (E). MoSE also demonstrates superior performance in terms of Hausdorff Distance (HD), with the lowest average HD of 17.0 mm, which is substantially lower than methods like D-Norm (75.8 mm) and CCSDG (24.4 mm). Specifically, MoSE achieves the lowest HD of 9.7 mm on AMOS (D) and 25.6 mm on CHAOS (C), while maintaining competitive performance on the other domains. DAPSAM, another ViT-based approach, shows better HD performance than most CNN-based methods, with an average HD of 27.6 mm.

For the ablation study, the MoSE w/o MoE model outperforms the baseline by 1.2\% in average Dice. More appealingly, our MoSE method further achieves a 3.2\% improvement in average Dice and a substantial reduction in average HD, dropping to 17.0 mm from 25.3 mm. As depicted in Fig.~\ref{seg_results}, the shape map accurately focuses on the target organ, providing both shape prior information and location information as a prompt for the subsequent segmentation. This guidance helps the model to better localize and segment the organ, thus improving both segmentation accuracy and generalization across unseen domains.

\subsection{Effectiveness of MoE Architecture.} In Fig.~\ref{hyper}(a), as the dictionary size increases from 128 to 256, the performance of the traditional method improves, reaching a peak at \(n=256\), with overfitting for larger $n$. In contrast, the MoSE architecture shows a continuous improvement in performance as the number of shape experts (denoted as \(n\)) increases. The best result achieved with MoSE is an average Dice of 91.4\% at \(n=1024\), demonstrating a clear improvement over the traditional method. This highlights the advantage of our MoSE approach, which effectively overcomes the trade-off between dictionary size and performance, providing superior results without the risk of overfitting observed in the traditional dictionary learning approach.

\subsection{Sensitivity Analysis of Hyperparameters.} As in Fig.~\ref{hyper}(b), when \(\beta\) is smaller than \(10^{-2}\), the penalty on the gating network is too weak, making it difficult to prevent the allocation of high weights to a few shape experts. As the model becomes overly reliant on a small subset of experts, it reduces generalization performance. Instead, larger \(\beta\) can result in an overly uniform selection of shape experts. In this case, the gating network fails to appropriately prioritize the most relevant experts for specific tasks, causing the model to lose its ability to exploit specialized knowledge from the shape experts effectively. In Fig.~\ref{hyper}(c), we found that setting \(T_{\text{warm-up}}=5\times10^{2}\) strikes a balance in training. When \(T_{\text{warm-up}}\) is smaller than \(5\times10^{2}\), the activation of shape experts is insufficient during the early stages of training. The Top-K selection can block gradients, preventing certain shape experts from learning useful shape information. Conversely, if \(T_{\text{warm-up}}\) is too large, the training process becomes slow in terms of shape experts' sparsification, which ultimately hampers the generalization.

\subsection{Number of Shape Experts and Top-K.}

\begin{figure}[t!]
\begin{center}
\includegraphics[width=1\linewidth]{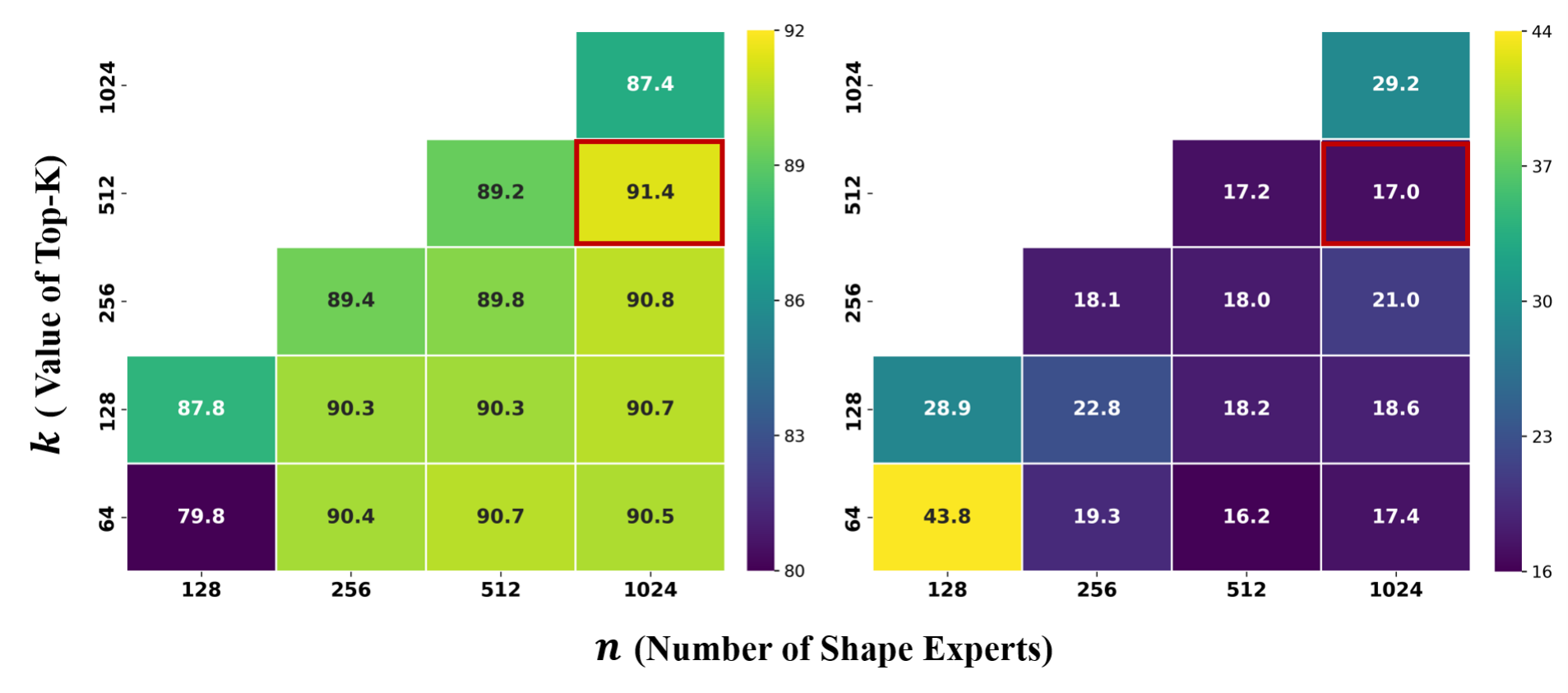}
\end{center}  
\caption{Dice (left) and HD (right) of different number of shape experts \( n \) and value of Top-K \( k \) on the generalization performance, calculated over all unseen domains. The diagonal (where \( n = k \)) represents the scenario without MoE.}
\label{nk_selection}
\end{figure}

These are crucial factors in our MoSE. As shown in Fig.~\ref{nk_selection}, the grid search range for \( n \) is set to 128, 256, 512, and 1024, while \( k \) ranges from 64 to 1024. Increasing \( n \) (i.e., storing more diverse shape experts) generally leads to better performance, as it allows the model to store a richer set of shape prior information, thereby enhancing its ability to generalize across unseen domains. Regarding the choice of \( k \), when \( n=256 \) or \( n=512 \), smaller values of \( k \) yield better performance, likely because the model benefits from focusing on fewer, more relevant shape experts, enhancing its generalization. However, when \( n=1024 \), larger values of \( k \) are more effective, as the increased number of shape experts requires a broader selection to effectively capture the most relevant shape priors.

\subsection{Visualization of Results on Unseen Domains}
As depicted in Figure~\ref{fig2}, MoSE achieves more accurate segmentation of the regions of interest (ROI), with its segmentation masks aligning more precisely with the anatomical structures compared to other methods, demonstrating its superior generalizable performance across unseen domains.

\subsection{Visualization of Shape Experts in MoSE}
As shown in Fig.~\ref{fig1}, the shape experts in our MoSE framework are visualized as heatmaps, each representing distinct shape priors learned by the MoSE. Each expert is represented as a heatmap, capturing distinct shape priors. The color intensity reflects the activation of each expert. These experts contribute to the segmentation task by dynamically selecting the most relevant priors at each pixel. The activation intensity of each expert varies, and the most relevant shape priors are dynamically selected for segmentation at each pixel.

\section{Conclusion}

We introduced MoSE, a novel end-to-end shape dictionary learning framework that leverages MoE principles to enhance SDG in medical image segmentation. Unlike conventional dictionary learning approaches that struggle with overfitting, and using suboptimal independently calculated shape dictionary, our method conceptualizes each dictionary atom as a distinct “shape expert” and employs a gating network to dynamically select the most relevant experts. By integrating the learned shape map as a prompt into the SAM, we further exploit the strong representation capabilities of large foundation models, achieving notable improvements on multiple unseen target domains. Future directions include extending MoSE to multi-class shape and 3D foundation segmentation models, as well as exploring other foundation models and prompt-engineering techniques to generalize beyond organ-level tasks. We believe our MoSE framework opens new avenues for domain-generalizable medical imaging by successfully merging shape prior learning with large-scale prompt-based segmentation models.

\section*{Acknowledgment}

This work is partially supported by NIH P41EB022544, R01CA290745, and R21EB034911.